%% file: IMUVIE_paper.tex
\title{IMUVIE: Pickup Timeline Action Localization via Motion Movies}

\pdfoutput=1

\documentclass[sigconf,nonacm,10pt]{acmart}

\acmConference{}{}{}

\setcopyright{none}
\renewcommand\footnotetextcopyrightpermission[1]{} 
\usepackage{graphicx}
\usepackage{wrapfig}
\usepackage{caption}
\usepackage{subcaption}
\usepackage{tabularx}
\usepackage{booktabs} 
\usepackage{float}
\usepackage{stfloats}
\usepackage{floatflt}
\usepackage{subcaption}  
\usepackage{tabularx, multirow, booktabs} 
\usepackage{float}
\usepackage{wrapfig}
\usepackage{amsmath,amssymb,amsfonts}
\usepackage{graphicx}
\usepackage{textcomp}
\usepackage{float}
\usepackage{multicol}
\usepackage{caption}
\usepackage{tabularx}
\usepackage{multirow}
\usepackage{booktabs,threeparttable}
\usepackage{array} 
\usepackage{booktabs}
\usepackage{tikz}
\usepackage{graphicx}    
\usepackage{wrapfig}   
\usepackage{caption}  
\usepackage{subcaption}  
\usepackage{wrapfig} 
\usepackage{multirow}
\usepackage{tabularx}
\usepackage{array} 
\usepackage{multirow} 
\usepackage{booktabs} 
\usepackage{etoolbox}
\makeatletter
\patchcmd{\thebibliography}{\chapter*{\bibname}}{\section*{\bibname}}{}{}
\patchcmd{\thebibliography}{\chapter*}{\section*}{}{}
\makeatother

\begin{document}

\begin{abstract}
Falls among seniors due to difficulties with tasks such as picking up objects pose significant health and safety risks, impacting quality of life and independence. Reliable, accessible assessment tools are critical for early intervention but often require costly clinic-based equipment and trained personnel, limiting their use in daily life. Existing wearable-based pickup measurement solutions address some needs but face limitations in generalizability.

\vspace{2mm}

We present IMUVIE, a wearable system that uses motion movies and a machine-learning model to automatically detect and measure pickup events, providing a practical solution for frequent monitoring. IMUVIE’s design principles—data normalization, occlusion handling, and streamlined visuals—enhance model performance and are adaptable to tasks beyond pickup classification.

\vspace{2mm}

In rigorous leave one subject out cross validation evaluations, IMUVIE achieves exceptional window level localization accuracy of 91-92\% for pickup action classification on 256,291 motion movie frame candidates while maintaining an event level recall of 97\% when evaluated on 129 pickup events. IMUVIE has strong generalization and performs well on unseen subjects. In an interview survey, IMUVIE demonstrated strong user interest and trust, with ease of use identified as the most critical factor for adoption. IMUVIE offers a practical, at-home solution for fall risk assessment, facilitating early detection of movement deterioration, and supporting safer, independent living for seniors.

\end{abstract}

\keywords{Health monitoring, Wearable technology, Automated pickup assessment, Inertial sensors, Timeline activity classification}



\author{\Large John Clapham\textsuperscript{$\diamond$}, Kenneth Koltermann\textsuperscript{$\diamond$}, Yanfu Zhang\textsuperscript{$\diamond$}, Yuming Sun\textsuperscript{$\mathsection$}, Evie N. Burnet\textsuperscript{$\star$}, Gang Zhou\textsuperscript{$\diamond$}}
\affiliation{
    \institution{\textsuperscript{$\diamond$}Dept. of Computer Science, \textsuperscript{$\mathsection$}Dept. of Mathematics, \textsuperscript{$\star$}Dept. of Kinesiology, William \& Mary}
    \city{Williamsburg}
    \state{VA}
    \country{USA}
}







\maketitle

\vspace{6mm}

\input{writing/introduction}
\input{writing/user_study}

\input{writing/system_design}

\input{writing/imu_movie_for_humans_and_ai}

\input{writing/top_modeling}

\input{writing/performance_evaluation}
\input{writing/discussion}
\input{writing/related_works}

\input{writing/conclusion}


\bibliographystyle{ACM-Reference-Format}
\bibliography{bib}

\end{document}

%% file: writing/introduction.tex
\vspace{-8mm}
\section{Introduction}\label{sec:intro}

Seniors often face challenges in picking up objects due to age-related movement ability deterioration, increasing the risk of falls and serious injuries such as bone fractures or head trauma~\cite{aziz2014distinguishing}. Falls have significant consequences, potentially leading to loss of independence and other socioeconomic and health-related effects~\cite{vaishya2020falls}. Nearly 3 million seniors visit the emergency room each year due to falls~\cite{CDCWISQARS}. 
Hospitalization may lead seniors to lose their independence, potentially resulting in a transition to assisted living. Falls are a significant risk, leading to fatalities at home, in nursing facilities, and in hospitals \cite{cross2022trends}. Even injuries like hip fractures can signal an approaching end-of-life event \cite{rapp2008hip, dubljanin2013does}.

\vspace{2mm}

A proactive healthcare approach can help prevent catastrophic and life-altering falls for seniors. Technology offers a viable avenue to monitor movement ability over time, allowing healthcare practitioners to assess fall risk and detect declines early for timely intervention. Monitoring movement ability through physical biomarkers, such as the capacity to pick up objects from the floor like a coaster or spoon, provides crucial insights that indicate the fall risk of an individual on any given day. This specific action, which requires bending, focus, and balance exemplifies a potential fall risk that could lead to serious injury on a hard surface such as a tile kitchen floor.
Early identification of declining pickup ability can prompt proactive treatments, such as physical therapy, to prevent injuries. Conversely, an improving pickup ability may be a sign that a physical therapy care plan is working.
Tests like the \textit{Berg Balance Test}~\cite{vieira2016prevention} can help identify issues early and allow for proactive intervention. 
The time taken to pick up an object from the floor is referred to as Time-of-Pickup (ToP). 

\vspace{2mm}

Current methods for assessing pickup ability over time require patients to visit clinics equipped with pressure-sensing mats like the ZenoMat~\cite{lynall2017reliability} or GaitMat~\cite{GaitMat}. Skilled practitioners analyze movement data, manually filter noise, and assign a ToP measurement and fall risk grade based on the subject's performance. This assessment is repeated periodically (e.g., monthly) at mat-equipped locations. The method relies on expensive hardware, takes up the valuable time of both patients and practitioners and lacks portability, requiring in-clinic assessments. This creates time and geographical limitations for patients, who have to schedule an appointment, travel to the clinic, perform testing, and then wait for results-- all while having to finance any hidden fees associated with the travel. To add more inconvenience for the senior, this process may take an inconvenient amount of time from initially scheduling the appointment to receiving notification of fall risk. Additionally, labeling ToP is labor-intensive, requiring the practitioner's prolonged attention. Time and money are required from the senior and the practitioner. Moreover, this entire clinical measurement process is not feasible to be conducted at frequent intervals (e.g., daily). There is a limit to the frequency that data can be collected, which limits the frequency that a fall risk can be calculated based on the changing ToP measurement. 
To solve these pain points, researchers have explored automatic measurement of ToP using wearable Inertial Measurement Unit (IMU) sensors. 
The state-of-the-art ToPick system \cite{clapham2024topick} employs ankle-worn sensors that transmit movement data to a mobile device, measuring ToP and allowing periodic monitoring of changes over time. Since ToP ability can indicate fall risk, ToPick could help identify increased fall risks early on. This system provides an alternative to clinic-based pickup assessments by enabling ToP and fall risk evaluation at home. However, ToPick uses a rule-based decision model, which often struggles to generalize to new subjects, especially outliers like those using walkers. The lack of generalizability is a barrier that needs to be solved.

\vspace{2mm}

To address these limitations, we ask the research question, 
\textbf{RQ1: ``How can we increase model generalizability for highly accurate ToP measurement for unseen subjects?"}

\vspace{2mm}



We present IMUVIE, a wearable solution that localizes and measures pickup events using motion movies and a vision-based machine-learning model. IMUVIE's model generalizes well and performs well on unseen data. IMUVIE can classify moments of pickup actions on unseen subjects that were never shown to the model training. Users perform pickup actions with motion sensors, which transmit data to an app that converts it into motion (IMU) movies, models the events, and delivers localized ToP measurements through a user-friendly interface. 
IMUVIE is guided by core design principles that enhance its ability at activity classification tasks, with applications broader than ToP measurement alone. The design involves eliminating redundant features that may hinder, rather than aid, the model's ability to classify accurately. Key design elements, such as plot size, axis labels, titles, legends, scale indexing, markers, anti-aliasing, interval selection, and sensor choice, are examined for their contribution to the spatial encoder's effectiveness. This optimized design maximizes our model's potential to successfully measure ToP.
To demonstrate our model's generalizability in classifying activity and measuring ToP on motion movie inputs, we evaluated it using pickup movement data from 33 senior participants collected in an IRB-approved user study. The results show high accuracy and outstanding recall at both event and window levels. Any areas of lower performance are intuitively explainable and offer clear avenues for future improvement. Survey interviews with 23 of these participants indicate strong acceptance and positive reception of our wearable ToP measurement system.
We make the following contributions in this paper:
\begin{enumerate}
    \item We introduce the IMUVIE pickup timeline action localization model, designed to identify pickup events from motion movie inputs. The model exhibits robust generalization and strong performance on unseen subjects, achieving high classification accuracy across 33 study participants at both the window and event levels. Notably, it achieves 91–92\% accuracy when evaluating 256,291 candidate motion movie frames. At the event level, the model demonstrates outstanding recall, reaching 97\% on 129 tested pickup events. This capability makes our wearable movement assessment system highly effective for measuring pickup ability.
    \vspace{2mm}
    \item We present a set of design principles to help create IMU movies for AI. The principles may maximize the performance of any vision model at any timeline activity classification task involving motion data even beyond pickup measurement.
    \vspace{2mm}
    \item We perform a user study on 33 senior individuals, and an interview survey on 23 seniors. The results indicate that seniors will adopt a system such as ours and that the highest factor that determines adoption will be the user-friendly design, above even the accuracy.
\end{enumerate}
 The remaining sections of this paper are organized as follows:  Section \ref{sec:user_study_data_collection} provides details on the user study conducted with 10 elderly individuals who participated in 38 pickup events. 
 Section \ref{sec:top_sys_design} outlines the design of IMUVIE. 
 Section \ref{sec:imu_movie} details the IMU Movie for Humans, and how we adapted its design to be tailored for AI.
 Section \ref{sec:top_modeling} details each part of the machine learning model, including the spatial and temporal encoders as well as the classifier. 
 In Section \ref{sec:performance_evaluation}, we present the rigorous leave one subject out cross validation performance evaluation. 
 Section \ref{sec:discussion} outlines future work inspired by our results.
 Section \ref{sec:related_works} discusses related works before we conclude in Section \ref{sec:conclusion}.

 

%% file: writing/user_study.tex


\section{User Study Data Collection}\label{sec:user_study_data_collection}


\subsection{Hardware, Data, \& Protocol}

    
    
    

\begin{wrapfigure}[9]{h}{0.1\textwidth}
\vspace{-4mm}
    \centering
    \begin{subfigure}[r]{\linewidth}
        \centering
        \vspace{-12mm}
        \includegraphics[width=0.95\linewidth,page=1,bb=0 0 822 822]{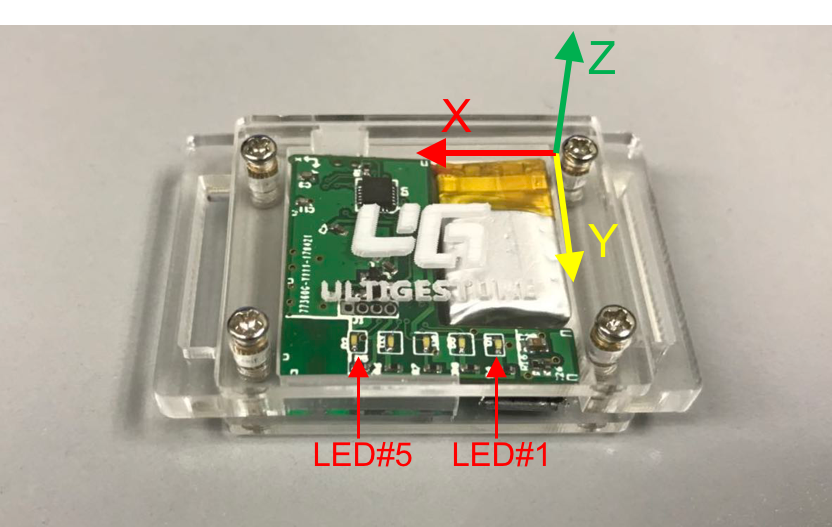}
        \caption{3D axis.}
        \label{fig:ug_axis}
    \end{subfigure}
    
    \vspace{2mm} 
    
    \begin{subfigure}[r]{\linewidth}
        \centering
        \vspace{-6mm}
        \includegraphics[width=0.95\linewidth,page=1,bb=0 0 625 625]{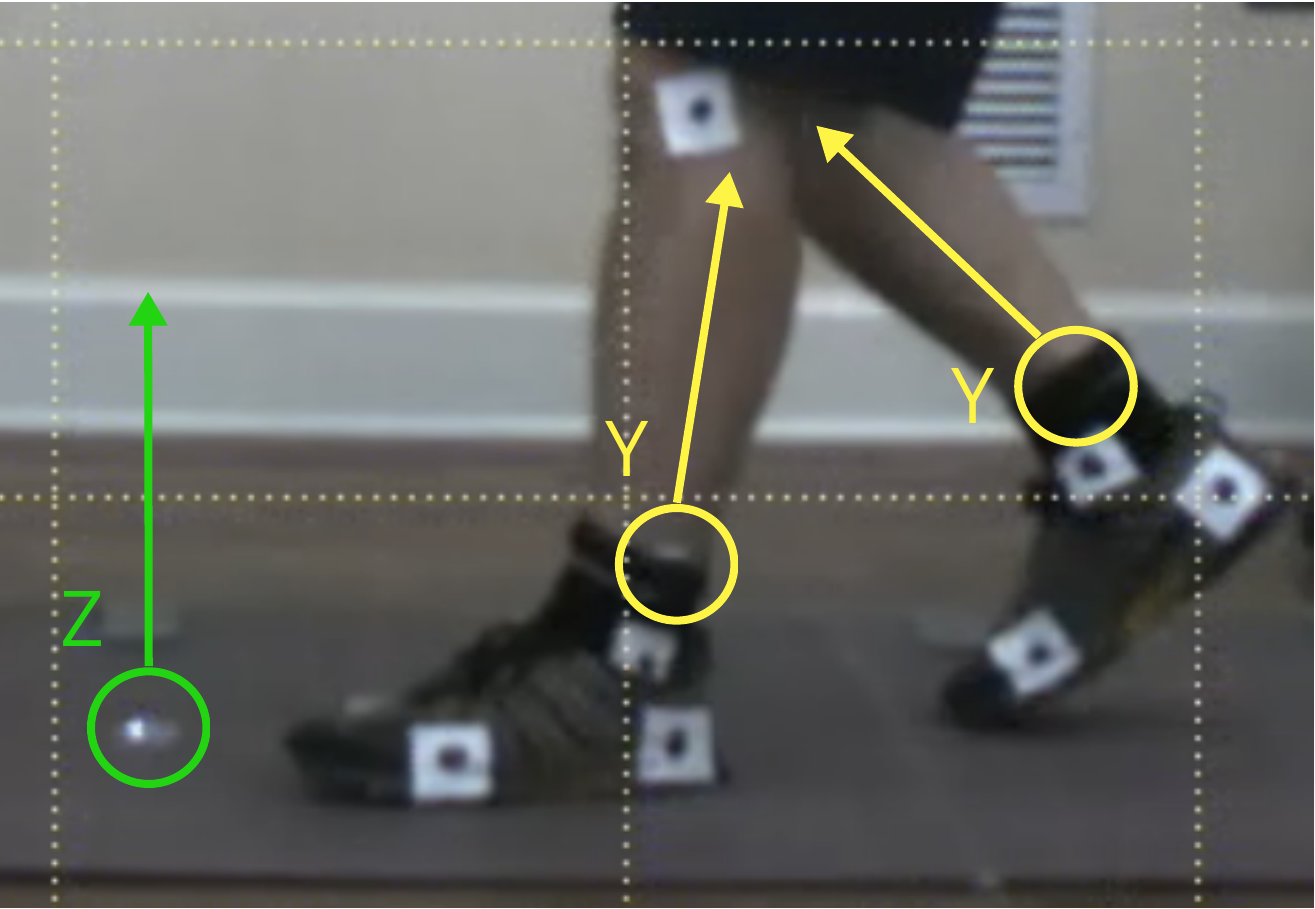}
        \caption{Setup.}
        \label{fig:setup}
    \end{subfigure}
    \vspace{-3mm}
    \caption{UG.}
    \label{fig:hardware}
\end{wrapfigure}

The Ultigesture (UG) wearable IMU sensor platform~\cite{zhao2019ultigesture} consists of a 3D gyroscope, accelerometer, and magnetometer (Figure \ref{fig:hardware}). Each device costs \$10 to manufacture and includes a Cortex-M4 processor with a BLE module. Before 2024, we used three UG devices: two ankle-mounted and one on the ground, as shown in Figure \ref{fig:setup}. 
\vspace{2mm}


The ankle sensors monitor foot movement. Figure \ref{fig:ug_axis} shows variations in the vertical axis based on sensor orientation. In 2024, we added a third IMU sensor to the chest to understand the anatomy of a pickup event in more detail. The UG devices sample IMU data at 100Hz, generating time-series data. 
Our model uses data from the ankle-mounted sensors only. Participants repeatedly walked, picked up an IMU device from the ground, and continued walking, before turning around and repeating the process to gather data across multiple pickup events.


\vspace{-1mm}

\subsection{Anatomy of a Pickup \& Ground Truths}

\begin{minipage}{0.5\linewidth}  
    \centering
    \includegraphics[width=\linewidth,page=1,bb=0 0 680 680]{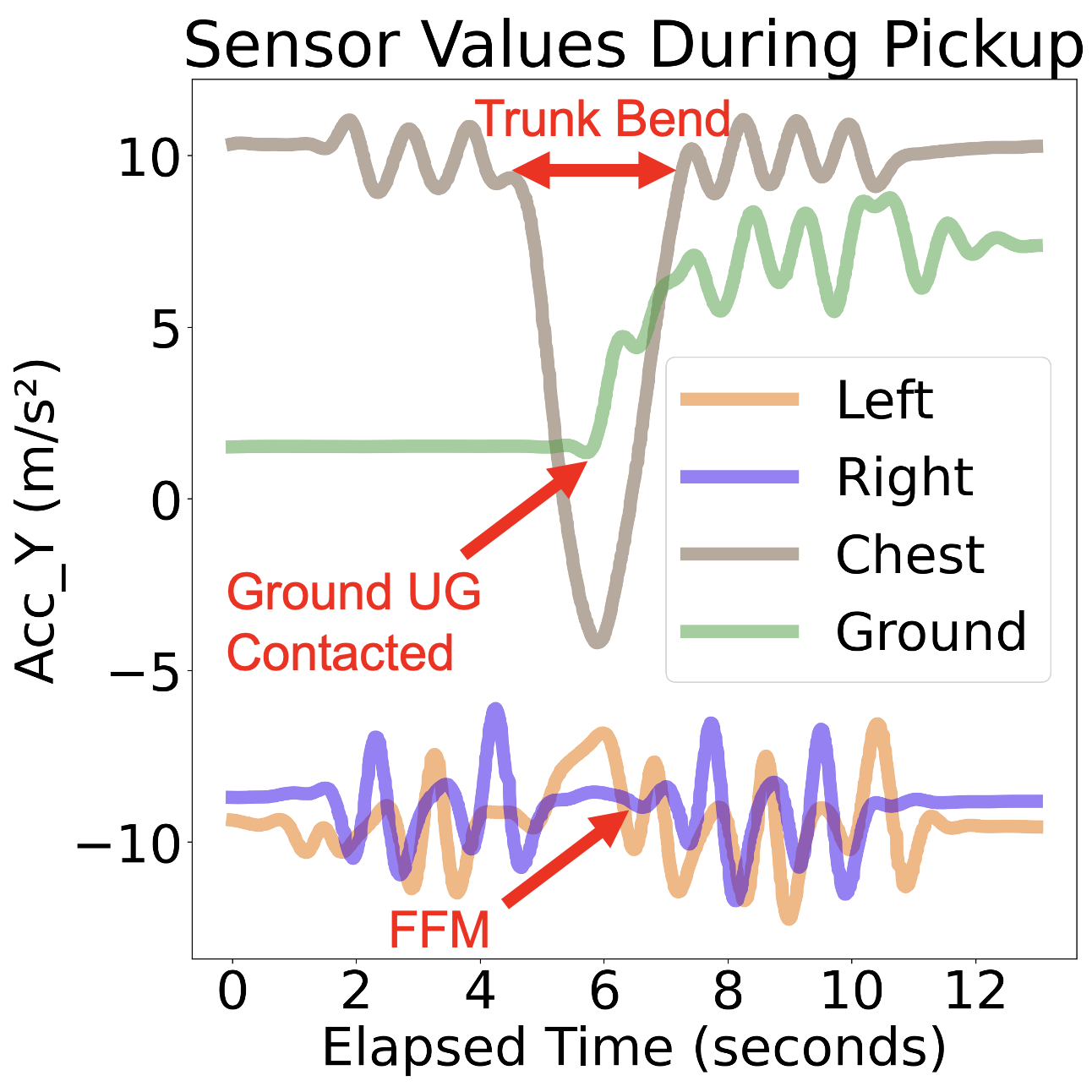}
    \captionof{figure}{ToP Anatomy.}
    \label{fig:anatomy}
\end{minipage}%
\hspace{0.05\linewidth}
\begin{minipage}{0.45\linewidth}  
    \vspace{2mm}
    Distinct patterns will indicate the following events. Figure \ref{fig:anatomy} shows three key moments of a pickup event: the start, contact, and first foot movement (FFM). The start indicates that the subject's trunk is bending towards the floor. The contact shows when the user interacts with the object on the ground. The FFM shows the end of the pickup event when the subject is now resuming their usual movement (e.g. walking).
\end{minipage}

\vspace{2mm}

\vspace{2mm}  

\hspace{-4mm}
\begin{minipage}{0.5\linewidth}  
Detection ground truth shown via the ground IMU device. When stationary, its vertical axis displays Earth's gravity (approximately 9.8 m/s²). Figure \ref{fig:contact} illustrates the the sensor's accelerometer values as a contact moment occurs. A spike signifies that the device has been ``contacted'' (i.e., physically moved) by the subject. 
A pickup event detection ground truth timestamp is marked at the spike in data. 
If any pickup event includes this contact timestamp, then the event is a true positive.
\end{minipage}
\hspace{0.05\linewidth}
\begin{minipage}{0.43\linewidth}  
    \centering
    \vspace{20mm}
    \includegraphics[width=\linewidth,page=1,bb=0 0 320 320]{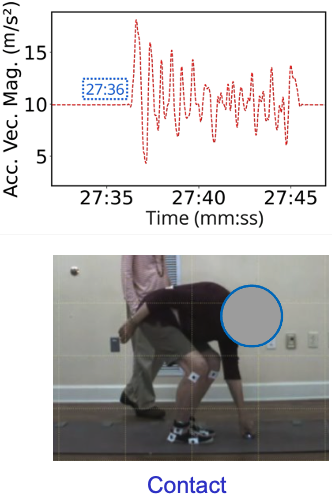}
    \captionof{figure}{Contact Moment.}
    \label{fig:contact}
\end{minipage}%

\vspace{2mm}


Traditional RGB imaging sensor (camera) videos serve as ground truth for determining the events' duration. These videos allow us to calculate the total duration of each pickup event by manually annotating the start and end times. To maintain consistency in labeling, we followed guidelines developed with an expert health practitioner, ensuring all subject videos were annotated according to the same standards.

\vspace{2mm}

\subsection{Subject Demographics \& Survey}

\vspace{2mm}

We collected time-series IMU data from 33 participants aged 74-99 with varied health conditions, stability, fall-risk, and pickup abilities. Details are in Table \ref{tab:demographics}. Data is anonymized, and collection follows an approved IRB protocol. The 10 subjects included in DS1 were collected before 2024. DS2 was collected in the Spring of 2024. The later group is also included in an interview survey which sought to understand perspectives on clinical methods, movement monitoring, and wearable technology among seniors. Table \ref{tab:survey_results} captures the resulting key observations. 

\begin{figure*}[htbp]
    \centering
    \includegraphics[width=\textwidth, bb=0 0 630 120, clip]{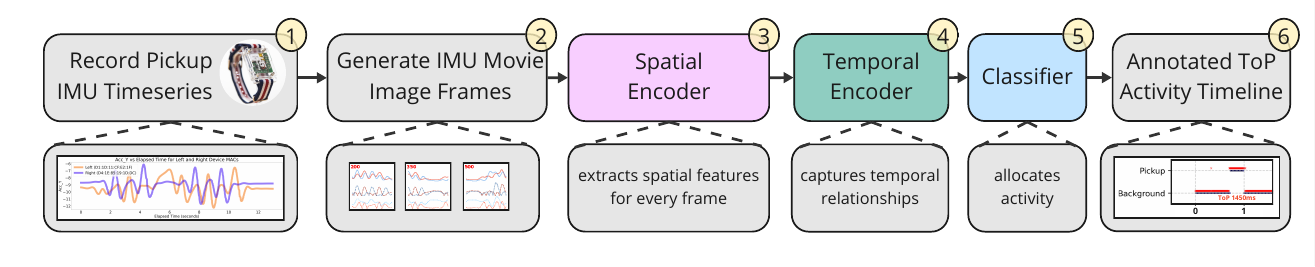}
    \vspace{-8mm}
    \caption{IMUVIE System Overview.}
    \vspace{-2mm}
    \label{fig:sys_overview_brief}
\end{figure*}

\begin{table}[h!]
\small
\centering
\renewcommand{\arraystretch}{1} 
\begin{tabular}{|>{\centering\arraybackslash}p{0.28\linewidth}|>{\raggedright\arraybackslash}p{0.25\linewidth}|>{\centering\arraybackslash}p{0.15\linewidth}|}
\hline
\textbf{Dataset} & \textbf{Characteristic} & \textbf{Detail} \\ \hline
\multirow{4}{=}{\centering DS1: Pre-2024} 
     & Subject Count & 10 \\ \cline{2-3} 
     & Age Range & 75-87 \\ \cline{2-3}
     & Pickups & 38 \\ \cline{2-3}
     & Surveys & 0 \\ \hline
\multirow{4}{=}{\centering DS2: Spring 2024} 
     & Subject Count & 23 \\ \cline{2-3} 
     & Age Range & 74-99 \\ \cline{2-3}
     & Pickups & 91 \\ \cline{2-3}
     & Surveys & 23 \\ \hline
\multirow{4}{=}{\centering Total} 
     & Subjects & 33 \\ \cline{2-3} 
     & Age Range & 74-99 \\ \cline{2-3} 
     & Pickups & 129 \\ \cline{2-3} 
     & Surveys & 23 \\ \hline 
\end{tabular}
\vspace{2mm}
\caption{Study Participants.}
\vspace{-2mm}
\label{tab:demographics}
\end{table}


The survey was divided into three main sections with a final open ended section. The first is aimed to assess seniors' opinions on current clinical methods and any pain points they may associate with the clinic. Participants were asked about the frequency and duration of their clinical health visits and how consistently they may follow advice from their medical practitioners. This section was intended to capture insights into any limitations or frustrations seniors experience with conventional clinical approaches to movement assessment and health monitoring. 

\vspace{2mm}

The second is focused on exploring seniors' attitudes toward home-based movement monitoring. Participants were asked about their interest in monitoring their health at home and their willingness to follow recommendations provided by AI-driven systems, though no specific solution details were shared. This section challenged the senior participants to provide valuable insights into the potential acceptance of home monitoring and AI recommendations without having any ideas about our solution. 

\vspace{2mm}

In the third, we aimed to assess seniors' experiences with technology and the feasibility of adopting wearable devices in their daily lives. Questions in this section helped us understand whether seniors would be comfortable incorporating wearable technology and whether it could realistically support their health and movement monitoring needs.The survey concluded with an open-ended section, inviting additional thoughts from participants. Most questions employed a Likert scale, though some open-ended questions invited open-ended feedback. Only the most prominent themes are presented in Figure \ref{tab:survey_results}, offering a focused summary of the survey findings. Overall, the results suggest that our proposed solution is likely to be positively received by seniors, indicating strong potential for adoption and engagement. 

\renewcommand{\arraystretch}{1.3} 

\begin{table}[H]
    \centering
    \small
    \vspace{2mm}
    \begin{tabular}{@{}p{0.19\linewidth} p{0.3\linewidth} p{0.3\linewidth}@{}}
        \toprule
        \textbf{Section} & \textbf{Key Observations} & \textbf{Survey Support} \\
        \midrule
        \textbf{Clinical Methods} 
        & Visits are time-consuming. High compliance with an expert's health advice. 
        & Average visit: 0.5–4 hours. Compliance: only 13\% unlikely to follow expert's advice. \\
        \midrule
        \textbf{Home \newline Monitoring}
        & High compliance with AI's analysis \& interest in home monitoring. Ease of use is important; ankle sensors preferred. 
        & 63\% likely to follow AI; 87\% interested in home use. 84\% ease of use; 2 don't like ankle sensors; 6 don't like chest sensors. \\
        \midrule
        \textbf{Technology Experience}
        & Most comfortable; Some used health tech. before. 
        & Only 22\% uncomfortable; 26\% have prior experience. \\
        \midrule
        \textbf{Additional Comments}
        & Seniors will try IMUVIE. Ease of use is important. Should be aware of history. 
        & ``Willing to try", ``Needs to be aware of prior (conditions) and easy to use". \\
        \bottomrule
    \end{tabular}
    \vspace{0.5mm}
    \caption{Survey Results Summary.}
    \label{tab:survey_results}
\end{table}

%% file: writing/system_design.tex
\section{IMUVIE System Overview}\label{sec:top_sys_design}

Figure \ref{fig:sys_overview_brief} shows an overview of the IMUVIE system. The process begins when the user records pickup movement data. Next, movie frame plots of this movement data are generated. These plots act as a sliding window over the time-series data, showing how the sensor values change over time as the user performs various activities. The frames are then processed by a spatial encoder, which captures spatial relationships between the plot frames. Following this, a temporal encoder captures the temporal relationships between frames. Finally, a classifier determines whether each frame represents a pickup event or background activity.

\begin{figure}[H]  
    \centering
    \includegraphics[width=0.5\textwidth,bb=0 0 8900 6100]{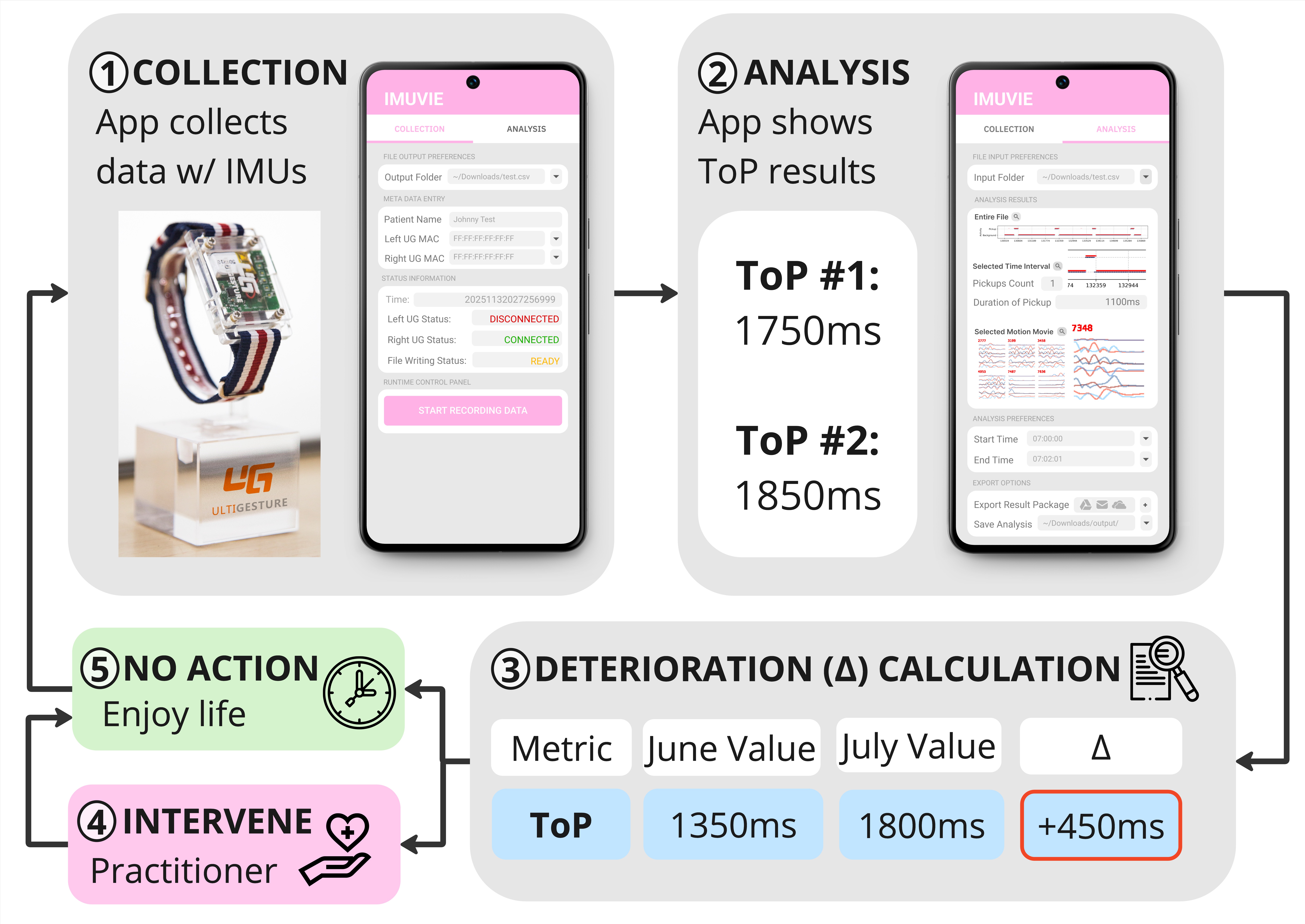} 
    \caption{IMUVIE in the Real World.}
    \label{fig:usage}
\end{figure}

Figure \ref{fig:usage} illustrates how IMUVIE can be used by individuals seeking to understand their pickup ability, assess fall risk, and monitor changes in movement ability over time.
In the first step, data is collected following the protocol outlined in the user study. The system is designed to allow unskilled users to gather data at any time or location. Seniors can collect data and assess fall risk during casual daily activities, such as walking their dogs or tidying their homes. 
Secondly, the mobile app receives the data from the sensors and processes it using AI. The analysis yields ToP measurements, which are then returned to the user, providing feedback on their pickup ability.
When used periodically, IMUVIE then calculates any deterioration in movement ability by comparing recent data to previous months' data. This comparison permits the detection of changes that signal an increased fall risk. 
An alert may be sent to the healthcare provider to signal that their expertise is needed.
After the intervention has been given, or a low fall risk is known, the seniors spend time enjoying life with a known and minimized fall risk. Keeping seniors at this stage is a primary goal of our system. This is where the fall risk is known and minimized. Users can focus on what is most important to them when spending their time between assessments.
The IMUVIE system usage cycle repeats after a designated period (e.g., daily, weekly, or monthly), allowing regular fall risk assessments to track gradual changes in movement ability over time.  
Changes in movement ability can indicate either deterioration or improvement. When used periodically, IMUVIE enables seniors and their practitioners to monitor movement deterioration that may signal an increased risk of falls. IMUVIE may also be used to demonstrate the effectiveness of physical therapy treatments, such as balance classes or home exercises, by generating quantitative measurements over time. These metrics provide a more precise gauge of treatment effectiveness than qualitative assessments alone and can show a gradually decreasing fall risk as a senior undergoes physical therapy.

\vspace{2mm}

\begin{figure}[H]  
    \centering
    \includegraphics[width=0.5\textwidth,bb=0 0 660 390]{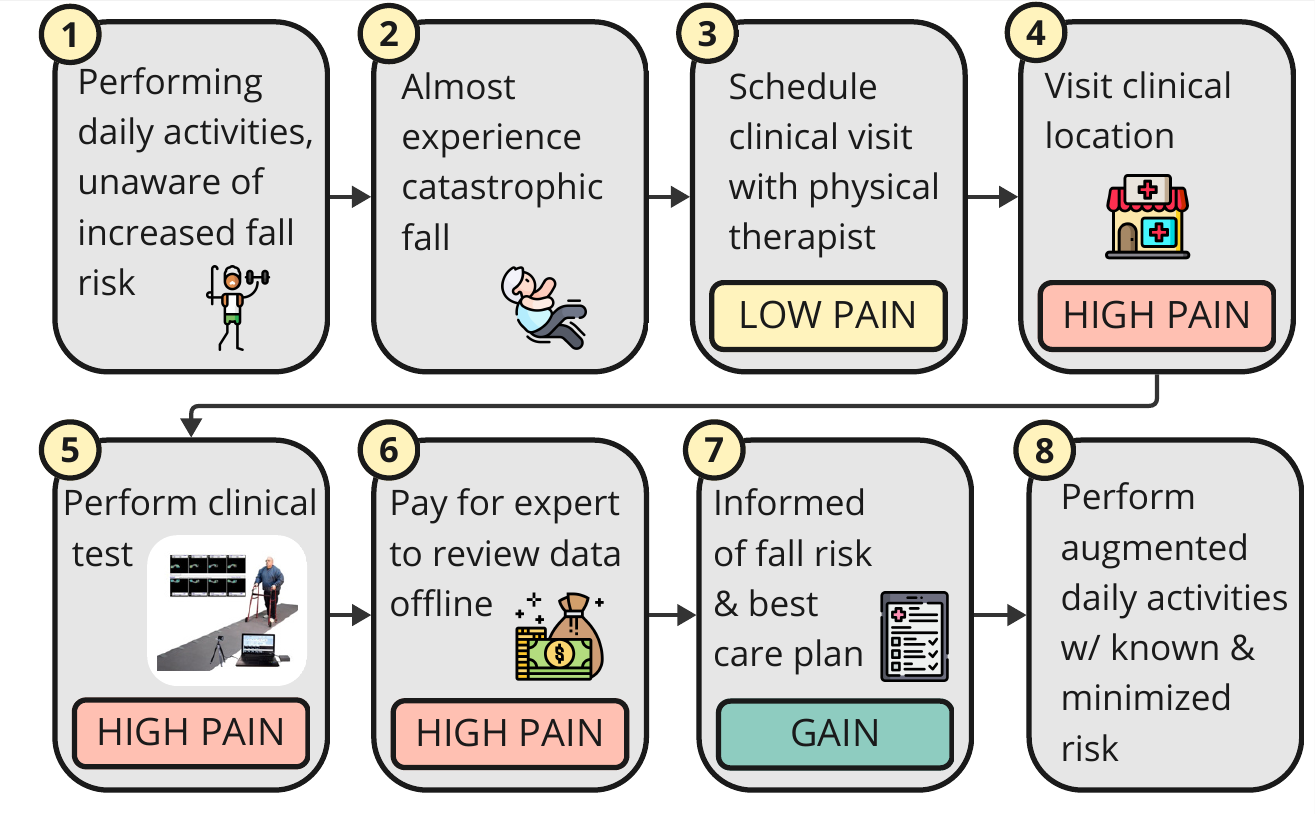}  
    \caption{Clinical Assessment Pain Points.}
    \label{fig:pain}
    \vspace{-2mm}
\end{figure}

This system provides value to two main user groups: seniors and practitioners. Currently, our focus is on the senior perspective. The IMUVIE usage cycle is designed to reduce four critical pain points (shown in steps 3, 4, 5, and 6 in Figure \ref{fig:pain}). The system can save time and money by removing all the pain points along the clinical movement assessment story.

%% file: writing/imu_movie_for_humans_and_ai.tex
\section{IMU Movie for Humans \& AI }\label{sec:imu_movie}

\begin{figure}[H]  
\vspace{-2mm}
    \centering
    \includegraphics[width=0.9\linewidth,page=1,bb=0 0 400 420]{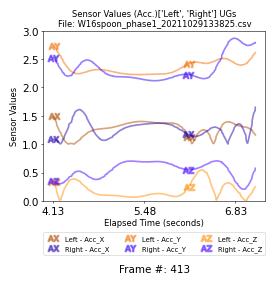}
        \caption{IMU Movie for Humans.}
        \label{fig:humanmovie}
\end{figure}

Our model analyzes a video and determines which frames belong to which activity class. The concept is straightforward: just as a human can watch the "IMU Movie" and pause to annotate activities, the model identifies activity classes in each frame.
Consider the example in Figure \ref{fig:humanmovie}. The leftmost timestamp corresponds to the frame number, and as time progresses along the x-axis, a human observer can see how the activity evolves. This visual process is similar to sliding a window along a time-series or sliding a time-series through a stationary window. We refer to this as the "IMU Movie." 

The IMU Movie runs at 100 frames per second (fps), meaning that a 10-second video contains 1,000 frames. A user can pause at any moment to assign an activity label to that particular frame. Each IMU movie frame corresponds to a 10-millisecond timestamp. In the example shown, the plot belongs to frame 413, which corresponds to a timestamp of 4130 ms. This may mark the start of a pickup event, with each subsequent frame being classified as part of the pickup event until the activity ends. The result is a contiguous array of IMU movie frames allocated to the pickup activity classification, while the rest of the IMU movie frames (before and after the event) are classified as background activity.

\vspace{2mm}


The IMU Movie is designed with humans in mind, incorporating visual elements like colors, markers, axes, titles, and legends for better comprehension. These elements, however, may not be relevant for an AI model during analysis. 

\vspace{2mm}

To address this unexplored challenge, we ask the sub-research question:

\vspace{2mm}

\textbf{RQ2: ``What set of design principles is essential for creating an optimal `Golden IMU Movie' input for AI? Specifically, an input that maximizes the model’s capability for accurate Timeline Activity Localization (TAL).''}

\vspace{2mm}

\noindent
\parbox{0.29\textwidth}{  

\vspace{2mm}
    To answer RQ2, we develop a set of design principles to optimize the IMU Movie specifically for AI:

    \begin{enumerate}
        \item Normalization Scale Indexing
        \item Occlusion Avoidance
        \item Anti-Aliasing Usage
        \item Interval Size Selection
        \item Sensor Selection 
        \item Plot Size \& Pixel Count
        \item Text on Axes, Legends, \& Titles
        \item Markers, Color \& Visual Style
    \end{enumerate}

    \vspace{2mm}

    Figure \ref{fig:aivshumans} shows the resulting ``Golden IMU Movie'' Input that is tailored for AI. This optimized input provides the best possible conditions for accurate timeline activity localization. Redundant features are removed. These principles can be adapted to any time-series problem, beyond just pickup activities from IMU data.

    \vspace{2mm}

} \hspace{6mm}
\parbox{0.14\textwidth}{  
    \includegraphics[width=\linewidth,bb=0 0 200 420]{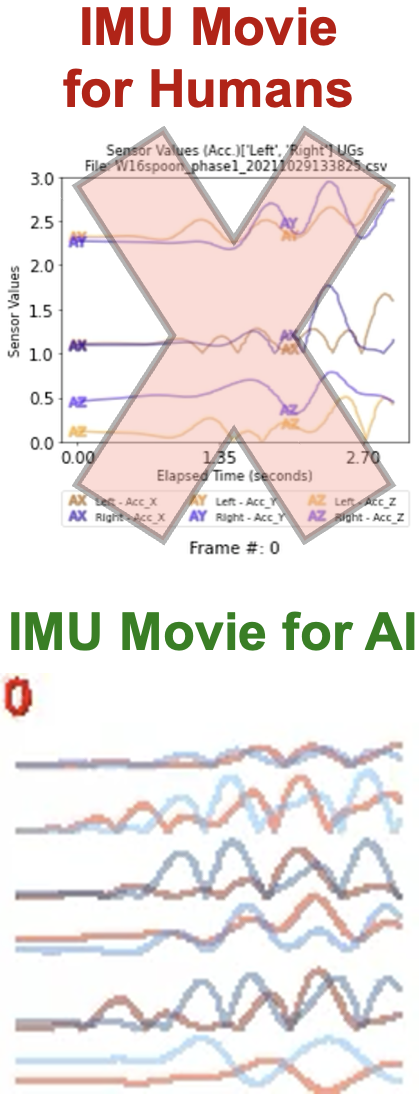}
    \captionof{figure}{\newline AI vs. Humans.}
    \label{fig:aivshumans}
}

\vspace{2mm}
\begin{wrapfigure}[10]{h}{0.2\textwidth}
\vspace{-12mm}
        \centering
        \includegraphics[width=\linewidth,page=1,bb=0 0 380 380]{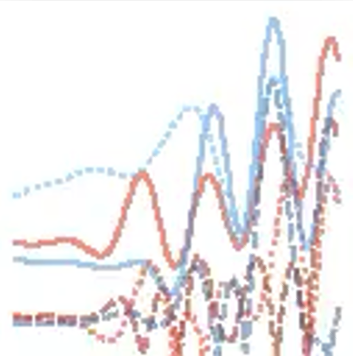}
        \caption{\newline Calibrated plot.}
        \label{fig:cal}
\end{wrapfigure}
\textbf{Normalization Scale Indexing.} Different sensors may have slight variations in calibration, leading to inconsistencies in the minimum and maximum readings between devices. These discrepancies can create challenges in recognizing patterns across sensors. To address this, we hypothesize that normalization can improve pattern recognition by standardizing the data. 
We normalize each sensor's readings based on its own minimum and maximum values, scaling them to range from 0 to 1. 
This approach ensures that each sensor's data is consistently represented, regardless of calibration differences.
After normalization, we obtain a plot, as shown in Figure \ref{fig:cal}, where all lines share the same y-axis range. However, this introduces a new problem: overlapping data lines can obstruct each other, making it difficult to analyze the visual information. To mitigate this, we must address occlusion.

\vspace{2mm}
\begin{wrapfigure}[12]{h}{0.2\textwidth}
        \centering
        \vspace{-2mm}
        \includegraphics[width=\linewidth,page=1,bb=0 0 180 190]{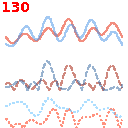}
        \caption{Occlusion-avoidance.}
        \label{fig:occlusion}
\end{wrapfigure}
\textbf{Occlusion Avoidance.} We hypothesize that plot lines are easier to analyze when they do not overlap and occlude one another. By minimizing the overlap between plot lines, as shown in Figure~\ref{fig:occlusion}, we aim to enhance both human and AI interpretability of the data. Visual separation allows key patterns to be observed without the distraction of multiple overlapping lines, which can obscure important information.
However, some degree of occlusion may be tolerable, or even beneficial, in specific contexts. For example, when two plot lines represent the same axis and sensor but are recorded from different feet, their partial overlap could indicate synchronization or other meaningful relationships between the signals. In such cases, careful design is needed to balance clarity with the potential value of overlapping information.

\begin{wrapfigure}[29]{h}{0.2\textwidth}
        \centering
        \includegraphics[width=\linewidth,page=1,bb=0 0 260 790]{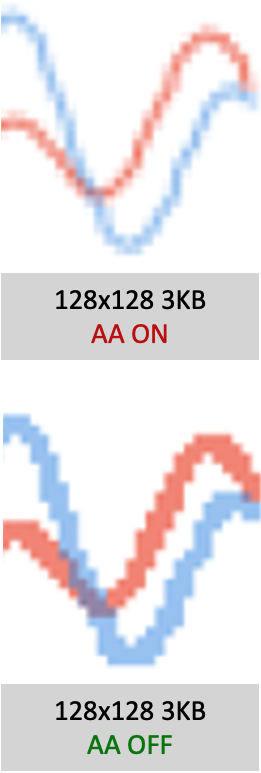}
        \caption{AA effects.}
        \label{fig:aa}
\end{wrapfigure}

\vspace{2mm}


\textbf{Anti-Aliasing Usage.} 
Common plotting libraries may introduce anti-aliasing to smooth the edges of shapes and text. This works well for standard-size images where we have hundreds of pixels of real estate. However, we hypothesize that anti-aliasing is not helpful for AI's analysis of these small images. When operating on tiny plots, anti-aliasing may introduce a blur, clarity loss, and distort the plot line's appearance, as is shown in Figure \ref{fig:aa}. All of these reduce image clarity and may hinder classification ability.
To avoid this side effect of blur and clarity loss when we reduced our plot sizes, we turned off this optional plot feature. The result is a plot without blur and clarity loss, ensuring sharper edges. Although this results in jagged edges, we suspect that this is not detrimental to the pattern recognition challenge faced by the AI model.


\begin{wrapfigure}[34]{h}{0.185\textwidth}
        \centering
        \includegraphics[width=\linewidth,page=1,bb=0 0 180 710]{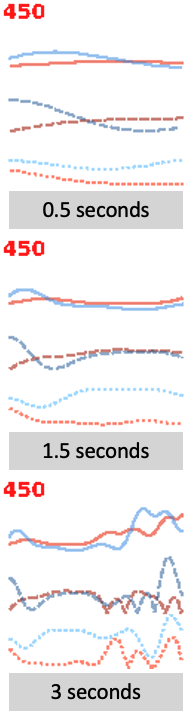}
        \caption{\newline Interval Sizes.}
        \label{fig:interval}
\end{wrapfigure}

\vspace{2mm}


\newpage

\textbf{Interval Size Selection.}
We hypothesize that a 3-second interval is ideal for spotting pickup events. Pickup actions typically last between 1 to 1.5 seconds, and having an additional 1.5 seconds allows us to capture the context after the pickup has been performed. This context can be valuable for understanding the current activity.
The example in Figure \ref{fig:interval} shows the approximate start of a pickup event. Which approach is easiest for a human to interpret? The answer depends on the specific domain problem being addressed. By considering intervals that provide sufficient context, we aim to improve both human and AI recognition of activity boundaries and transitions.



\vspace{2mm}

\textbf{Sensor Selection.}
We hypothesize that just the accelerometer may not be enough to distinguish between pickup events and turn events.
Both the gyroscope and accelerometer should offer superior features to the model compared to just the accelerometer. 
This will help activity classification.
This should reduce the false positives since it can distinguish more detail between pickups and turn events.
We found that we can reduce the false positive rate by adding gyroscope information for 6 plot rows with 12 lines.

\vspace{2mm}

\textbf{Plot Size \& Pixel Count.} The model trains on sequences of images, which can consume significant memory resources, even on servers equipped with high-performance GPUs. To ensure smooth training and classification, we must consider performance optimizations.
We hypothesize that reducing the size of each frame will not hinder classification accuracy and may even enhance it. Optimizing the plot size offers benefits beyond avoiding resource constraints: for example, it allows us to increase sequence length or add more model complexity without sacrificing performance. To achieve this, we reduce the absolute size of each IMU Movie frame to 64x64 pixels. As a result, each frame is just 0.5-3KB in size, depending on the content that is captured, making it more efficient for processing while retaining essential information for accurate activity classification.


\vspace{2mm}

\textbf{Text on Axes, Legends, \& Titles.} We hypothesize that text elements, such as axis labels and legends, may introduce noise for the model and may not aid in classification. In fact, they could even hinder the model's classification ability. We remove text to we can also reduce the plot size without having to consider the visual appearance of the text elements.
To accommodate both human understanding and AI optimization, we create two versions of the IMU Movie for any given input time series. The first is a "debug" plot, which includes text elements and is intended for use by human engineers to understand and modify the system. This debug plot, shown earlier in Figure \ref{fig:humanmovie}, is also referred to as the IMU Movie for humans. The second version is a "production" plot, which is stripped of redundant text elements and serves as the input for the model.

\vspace{2mm}

\textbf{Markers and Color.}
Certain visual styling options, such as special colors and floating markers, may help humans identify patterns more easily. However, this may not be the case for AI models. We hypothesize that the use of special colors, floating markers, and other visual styling enhancements may only introduce noise for the model and provide little to no benefit for classification. In fact, these visual elements could potentially hinder the model's ability to accurately classify activities. A makeover using simple visual styling allows us to create a cleaner input that focuses solely on the relevant data and enhances the model's ability to learn effectively.

\vspace{2mm}

\begin{wrapfigure}[14]{h}{0.25\textwidth}
        \centering
        \includegraphics[width=\linewidth,page=1,bb=0 0 180 180]{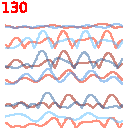}
        \caption{\newline Golden IMU Movie.}
        \label{fig:goldenfleece}
\end{wrapfigure}

\textbf{The Golden IMU Movie.}  Figure~\ref{fig:goldenfleece} shows the resulting movie frame design that follows the set of design principles that maximize the model’s ability for TAL as desired by RQ2. 

%% file: writing/top_modeling.tex
\vspace{-2mm}
\section{IMUVIE TOP Modeling}\label{sec:top_modeling}


The following section introduces the spatial encoder, the temporal encoder, and the classifier. This modeling diagram details how we get from the collection to the analysis stages of the IMUVIE usage cycle discussed earlier. Each of these components plays a key role in processing and classifying the sensor data collected from IMU devices during pickup events. The spatial encoder identifies patterns in individual frames, the temporal encoder captures changes over time, and the classifier assigns labels to each frame, distinguishing between pickup events and background activities.
\vspace{2mm}

 The model classifies actions within an input video of sequential movie frames \( f(X) \rightarrow (\hat{Y}) \), where function \( f \) is a machine learning model. 
We use notation consistent with ActionFormer~\cite{zhang2022actionformer}. The input video \( X = \{ x_1, x_2, \dots, x_T \} \) serves as the input to the model, which outputs a corresponding set of estimated classification action labels \( \hat{Y} = \{ \hat{y}_1, \hat{y}_2, \dots, \hat{y}_n \} \). The set \( X \) has a ground truth label set \( Y = \{ y_1, y_2, \dots, y_n \} \) that defines the action boundaries for each frame. Each element in \( Y \) and \( \hat{Y} \) is a tuple \( y_i = (s_i, e_i, a_i) \), where \( s_i \) is the start time, \( e_i \) is the end time, and \( a_i \) is the action.

\begin{wrapfigure}[14]{R}{0.16\textwidth}
\vspace{-16mm}
        \centering
        \includegraphics[width=\linewidth,page=1,bb=0 0 4700 10000]{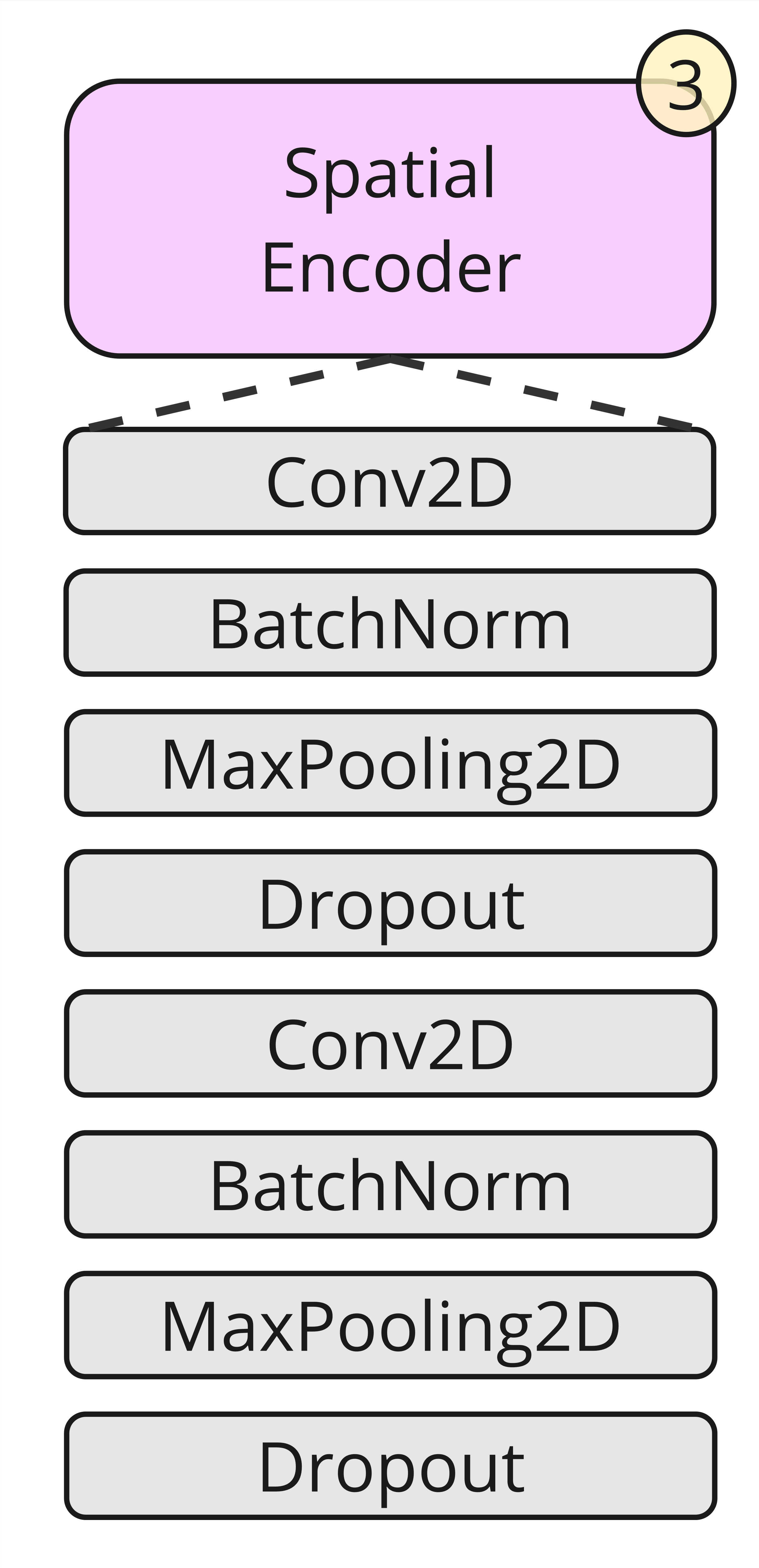}
        \vspace{-8mm}
        \caption{\newline Spatial Encoder.}
        \label{fig:spatial}
        \vspace{-3mm}
\end{wrapfigure}

\subsection{Spatial Encoder}

The Spatial Encoder is designed to learn spatial features from the IMU Movie Input frames and is shown in Figure \ref{fig:spatial}. It focuses specifically on recognizing patterns that are indicative of a pickup event. Given the input data, which consists of line plots $X = \{ x_1, x_2, \dots, x_T \}$. containing the movement information extracted from ankle-mounted IMU sensors. The encoder maps $X$ into $E(X) = E(x_1), E(x_2), ..., E(x_T)$ where $\mathbb{E}(\mathbf{x}_i) \in \mathbb{R}^D$.
\vspace{2mm}

The encoder uses only two convolutional layers to extract relevant spatial information. 
Each 2D CNN layer is encased in a TimeDistributed wrapper, allowing convolution to be applied to each frame individually within the sequence. 
We selected a 3x3 kernel size in the first layer and 10x10 in the second layer. We choose ReLU as the activation function. The input shape is defined as (sequence length, size, size, 3), where the sequence length is 10, the size is 64x64 (the Golden IMU movie frame size). After each CNN layer, we add batch normalization, max pooling, and dropout to avoid overfitting.

\vspace{2mm}

The encoder architecture is different from a traditional CNN for timeline image classification.
Our minimalist spatial encoder is thoughtfully constructed for the IMU activity classification task at hand. The use of fewer convolutional layers with fewer units has several advantages. One is the reduced computational cost and faster training times. It also helps mitigate the risk of overfitting. In contrast, traditional vision networks applied to real-world images (e.g. derived from still photos or movie frames) require a deeper architecture with more layers and units to capture complex visual patterns, such as textures, shapes, and objects. We do not need to consider this complexity for the line plot frame images. 
This lightweight architecture is effective because the patterns present in IMU data are relatively simple compared to typical visual features in real-world images. Pickup events are often distinguishable from the background as distinct line patterns will occur. 
These two convolutional layers are sufficient to capture these recurring features. Additional layers may even add unnecessary complexity, potentially leading to overfitting without significant performance improvement.


\subsection{Temporal Encoder}
\begin{wrapfigure}[13]{R}{0.16\textwidth}
        \centering
        \vspace{-7mm}
        \includegraphics[width=\linewidth,page=1,bb=0 0 270 400]{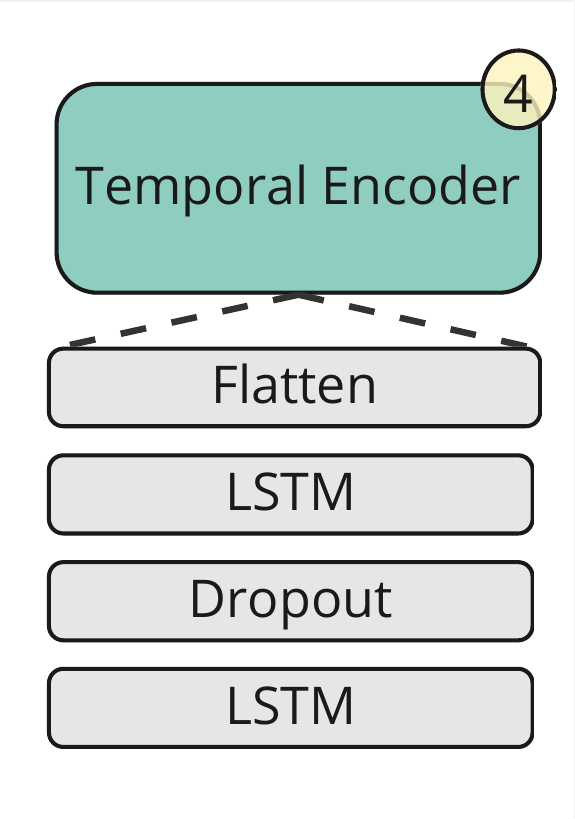}
        \vspace{-4mm}
        \caption{Temporal Encoder.}
        \label{fig:temporal}
        \vspace{-3mm}
\end{wrapfigure}

The Temporal Encoder learns temporal features from the IMU Movie Input frames by recognizing sequences of line patterns that indicate a pickup event. It is shown in Figure \ref{fig:temporal}. 
This component helps the model classify many sequential frames of a pickup action, resulting in a more reliable classification than when looking at just a single frame to classify an activity. Pickup events are indicated by line plot pattern changes that span over consecutive frames, and the Temporal Encoder is perfect for identifying these changes.

\vspace{2mm}

We train the model using sequences of frames, and not just single IMU movie frames. Just as a human observer would find it difficult to classify an action from a single frame, the model needs temporal context to identify pickup events. 
\begin{wrapfigure}[11]{h}{0.16\textwidth}
\vspace{-2mm}
        \centering
        \vspace{-8mm}
        \includegraphics[width=\linewidth,page=1,bb=0 0 320 400]{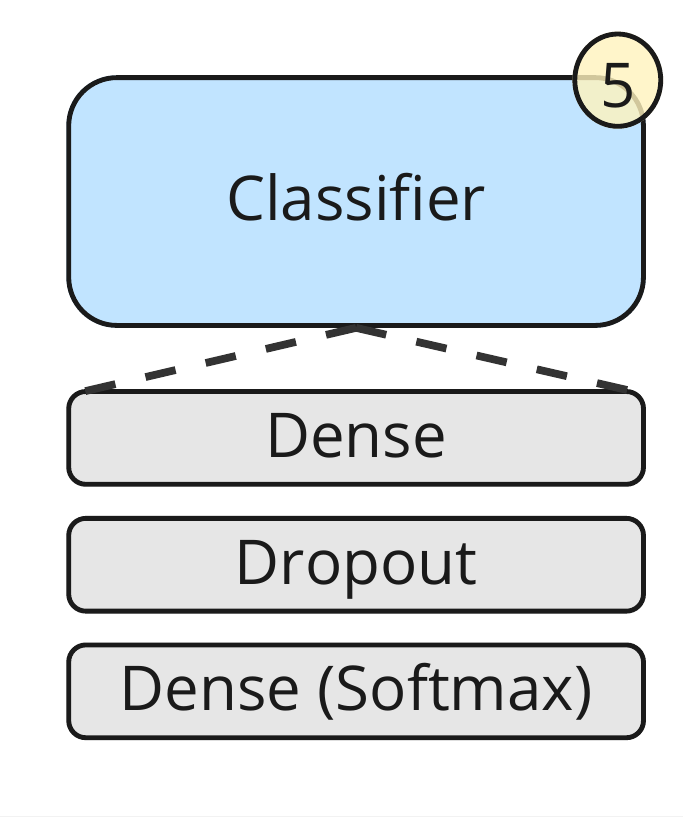}
        \caption{\newline Binary Classifier.}
        \label{fig:classifier}
        \vspace{-3mm}
\end{wrapfigure}
Humans use video, not a single image, to understand actions, and the AI should mimic this by using the same type of temporal information for accurate classification. 
We use a sequence length of 10 frames, which corresponds to 100 ms of data. We choose a stride length of one. This window length helps the encoder capture the temporal relationships needed to recognize pickup events.

\vspace{-1mm}
\subsection{Classifier}

\begin{figure*}[htbp]
    \centering
    \includegraphics[width=\textwidth,page=1,bb=0 0 15000 3000]{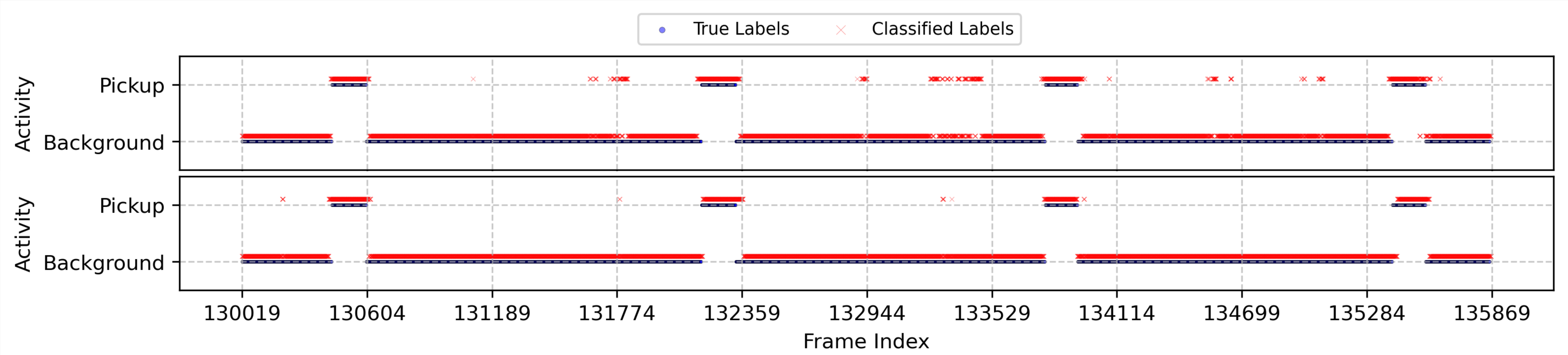}
    \caption{Four Pickup Events. (Top) False Positives Discovered. (Bottom) False Positives Mitigated.}
    \label{fig:windowlevelexample}
\end{figure*}

The classifier determines whether the current frame represents a pickup event or a background activity. It is shown in Figure \ref{fig:classifier}. It produces the output set \( \hat{Y} = \{ \hat{y}_1, \hat{y}_2, \dots, \hat{y}_n \} \) where each \(\hat{y}_i\) corresponds to a unique input frame \(x_i\) and consists of the tuple \( y_i = (s_i, e_i, a_i) \). The activity set can be visualized on a time-series and is shown in Figure \ref{fig:windowlevelexample}.




\vspace{2mm}

After the Spatial and Temporal Encoders have extracted their features, the classifier uses the information to decide the current activity for each frame. We chose the softmax activation function since this gives us the ability to label more classes than pickup alone. We may also add additional labels (e.g., turning or sitting down). 

\vspace{2mm}

This binary classification step is the final step for distinguishing between pickup actions and background frames. The model assigns a label to every frame of sensor data: either background or pickup. 
This frame-by-frame labeling provides a detailed temporal understanding of when pickup actions occur.

\vspace{2mm}

Each frame's classification result corresponds to a specific timestamp. We have a classification every 10 ms since our sensors operate at 100 Hz and our IMU Movies are 100 FPS. 
For example, frame 0 may represent a time span of three seconds, but its classification result is associated with the timestamp 0 ms. Similarly, frame 100 may span from 1000 ms to 4000 ms, and its classification result will correspond to the timestamp 1000 ms. This ensures that each classification is precisely linked to a specific point in time, allowing for an accurate temporal representation of pickup events.

%% file: writing/performance_evaluation.tex
\vspace{-1mm}

\section{Performance Evaluation}\label{sec:performance_evaluation}

\subsection{Settings} 
We conduct two levels of evaluation: the first at a window level and the second at a pickup event level.
The dataset includes 256,291 IMU movie frames, equivalent to over 42 minutes of activity.

\vspace{2mm}

\begin{wrapfigure}[15]{R}{0.28\textwidth}
        \centering
        \vspace{-3mm}
        \includegraphics[width=\linewidth,page=1,bb=0 0 8000 6500]{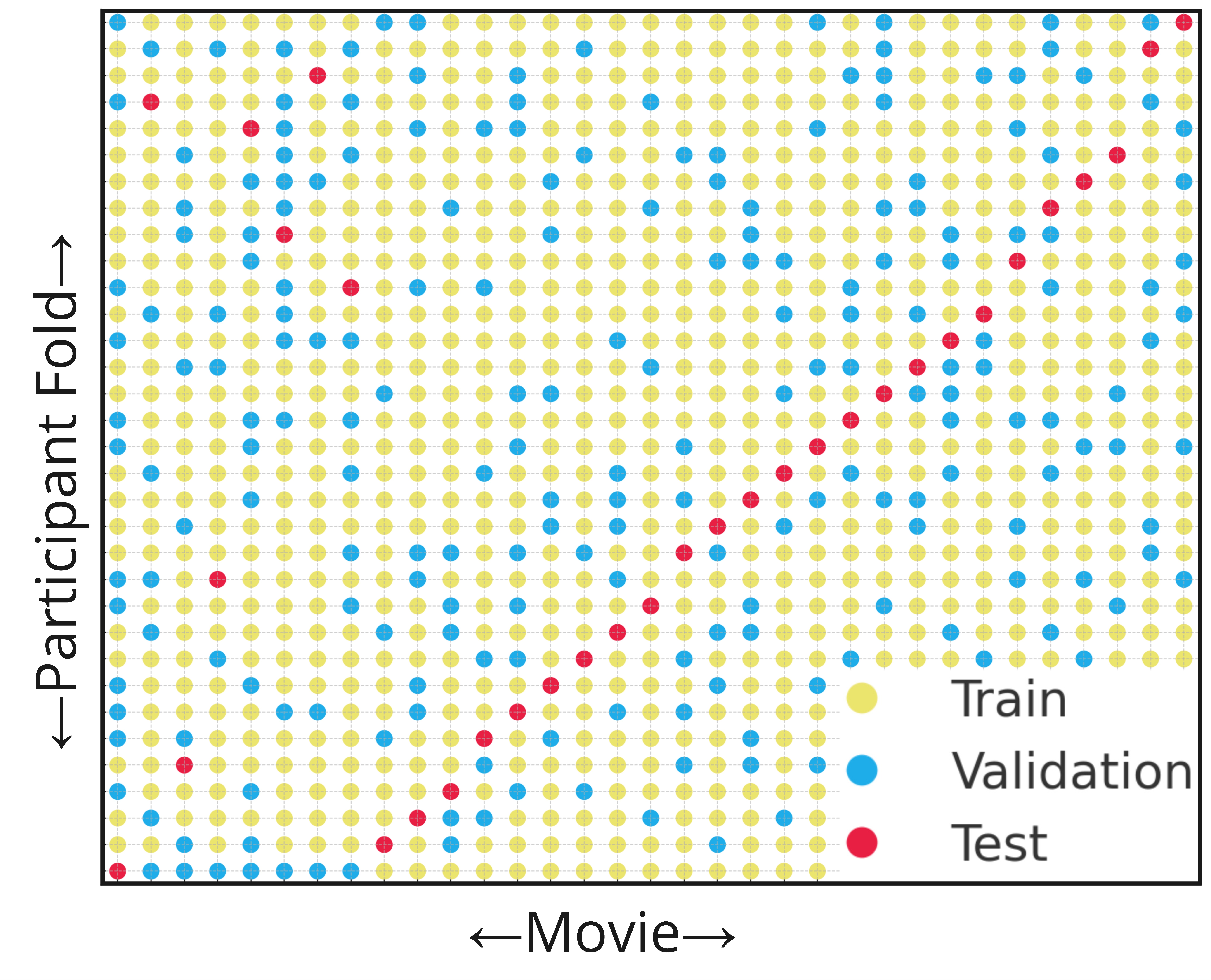}
        \caption{LOSOCV Set Selection for 33 Folds (Round One).}
        \label{fig:LOSOCVsetselection}
        \vspace{-2mm}
\end{wrapfigure}

We evaluate the performance at each 10 ms frame granularity level for each of the 256 thousand IMU movie frames. We perform leave one subject out cross validation (LOSOCV) on the 33 study participants. The training set consists of 25 subjects, while the test set is comprised of data from one subject. The validation set includes seven movies. There is no data overlap between sets within each fold. This ensures we have no data leakage during the training and evaluation process. The real set selection for the first round is shown in Figure \ref{fig:LOSOCVsetselection}. We conduct these 33 experiments and synthesize the results for a final aggregated performance. This test challenges the model to generalize from training data to perform well on unseen data. If successful, we will have answered our research question RQ1. The allocation of the test and training set is random every time we replicate the evaluation. We expect variance in our results. This random selection is important to prove that our results are not tied to a lucky set selection.

\vspace{2mm}

Event level evaluations are also important because they demonstrate our ability to detect a pickup within untrimmed movement data. We count all contiguous sequences of classified frames as an event classification. For example, if we have 100 contiguous frames classified as a pickup, that counts as one pickup event lasting 1000 ms. Even one disconnected movie frame allocated to just one 10 ms time interval may be an event. There are a total of 130 pickup events performed across the 33 subjects.

\vspace{-2mm}

\subsection{Window Level}

Our leave one subject out cross validation on 256,291 motion movie frames spanning 33 participants demonstrates the IMUVIE model’s strong generalizability and high accuracy in measuring ToP. As shown in Table \ref{tab:cross_validation_results}, the variance in results from multiple evaluations reflects the robustness of our approach, providing reliable results that are not subject to luck.



\begin{figure*}[htbp]
    \centering
    \includegraphics[width=\textwidth,page=1,bb=0 0 13900 3500]{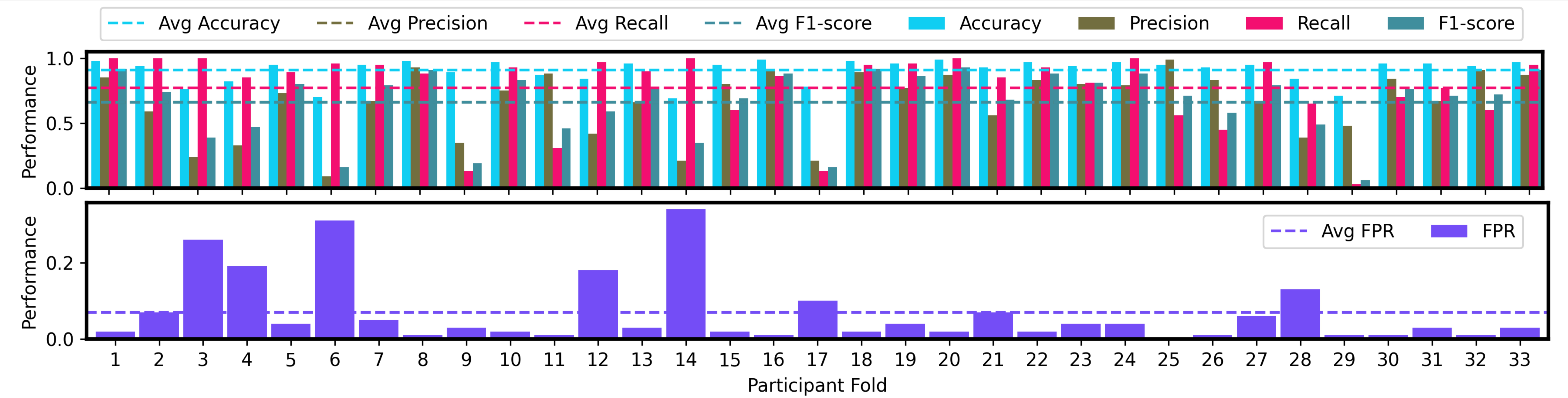}
    \caption{Leave-One-Subject-Out Window Level Test Performance Across 33 Participants.}
  \label{fig:losocv}
\end{figure*}

\begin{table}[h!]
    \small
    \centering
    \begin{tabular}{|p{0.2\linewidth}|p{0.2\linewidth}|}
        \hline
        \textbf{Metric} & \textbf{Mean} \\ \hline
        Accuracy & 91-92\%  \\ \hline
        Precision & 66-70\%  \\ \hline
        Recall & 77-78\%  \\ \hline
        F1 & 0.66-0.70 \\ \hline
        FPR & 5-7\%  \\ \hline
    \end{tabular}
    \vspace{4mm}
    \caption{Averaged LOSOCV Results.}
    \vspace{-6mm}
    \label{tab:cross_validation_results}
\end{table}

\textbf{False Positive Classification Rate.} The lower portion of Figure \ref{fig:losocv} illustrates the false positive rate for each fold in our leave-one-subject-out evaluation across 33 participant folds. Overall, we observe a low false positive rate across all folds, with an average of only 7\%. A comparison between the upper and lower sections of the figure indicates that a higher false positive rate correlates with lower precision. For instance, in participant fold 14, the false positive rate is relatively high, while precision is low. Further reducing these false positives is feasible and could be achieved by optimizing the model, which we consider a natural extension of this work.
\vspace{2mm}

Figure \ref{fig:windowlevelexample} illustrates the effect of one method that we have already incorporated to reduce falsely classified pickup frame windows. In the initial version of IMUVIE, we provided only accelerometer data to the model. Accelerometer data is highly effective at detecting foot motion, such as walking, jumping, or running, but it struggles to differentiate pickups from some other event types with similar acceleration activity. For instance, a subject slowing down before picking up an object might exhibit movement patterns similar to someone pausing before turning around at the end of a walkway. The accelerometer alone cannot reliably distinguish between genuine pickup events and turning events involving a pause.
To address this limitation, we incorporated gyroscope data into the motion movies. This addition significantly reduced false positive frame classifications and improved average performance metrics across all LOSOCV participant folds. 
Future work could investigate handcrafted features designed specifically to distinguish between turns and pickup events, further enhancing window-level performance.

\vspace{2mm}

\textbf{ToP Detection Performance.}
Our model demonstrates exceptional recall, accurately detecting nearly all 10ms granularity windows labeled as pickup events. When a 10ms window's movie frame corresponds to a pickup event, our model reliably and correctly classifies it. The high level of average accuracy across unseen subjects demonstrates that we have answered our initial research question, which demanded a generalizable model capable of accurate ToP measurement. Our design principles proposed in response to RQ2 have helped our model's ability to perform accurate TAL.

\vspace{2mm}

\textbf{Balancing Window Level Performance Tradeoffs. } Our goal is to balance performance metrics to maximize utility for the end users of our system. On the one hand, we must minimize the misclassification of genuine pickup movie windows. On the other, we must reduce false positive classifications for these windows. Achieving this balance involves fine-tuning the model by experimenting with factors such as complexity, architecture, activation functions, units, filter size, and other parameters.
The results of our extensive LOSOCV experiments demonstrate the effectiveness of our chosen model design. We hypothesize that further parameter adjustments may provide only incremental improvements and would enhance some metrics at the cost of others. 


\vspace{-4mm}


\subsection{Event Level}

To assess performance at the level of individual pickup events, we evaluate classification ability across participants. In total, there are 129 pickup events to classify among the 33 subjects: 31 subjects performed 4 pickups, one performed 3, and one performed 2.
At this stage of the evaluation, we focus on whether the pickup event is detected at all, rather than pinpointing its exact start and end times. While window-level performance is better suited to assessing precise accuracy in ToP measurement, this event-level evaluation examines IMUVIE's ability to determine whether an event has occurred, demonstrating model generalizability. Both evaluations are critical for addressing RQ1, showing that we have developed a generalizable and highly accurate ToP measurement system that performs well on unseen subjects.

\vspace{2mm}

\begin{figure}[H]  
\vspace{-2mm}
    \centering
    \includegraphics[width=0.99\linewidth,page=1,bb=0 0 7700 6000]{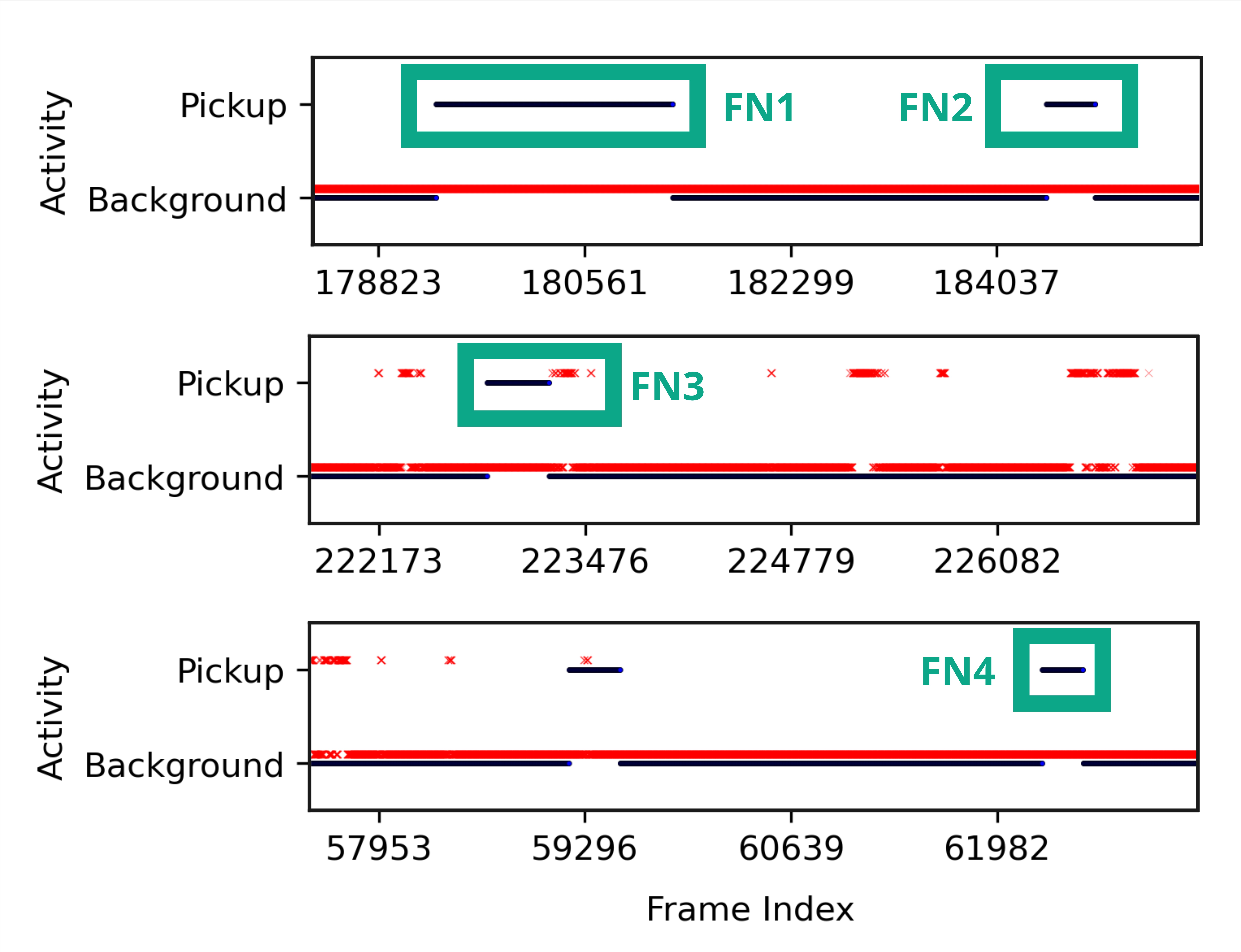}
        \caption{Four False Negatives.}
        \label{fig:fourfalsenegatives}
    \vspace{-2mm}
\end{figure}

\textbf{True Classification Rate.} Out of the 129 pickup events, we correctly classified 125, achieving a recall of 97\%. Figure \ref{fig:fourfalsenegatives} shows the four false-negative event-level classifications. The final pickup for subject fold 9 was missed (FN4), as was the third pickup for participant fold 17 (FN3), along with the third and fourth pickups for participant fold 29 (FN1 and FN2). Cross-referencing these results with Figure \ref{fig:losocv} shows that these three folds had below-average performance, particularly in recall. Further investigation suggests that these participants exhibited unusual pickup styles, indicating that they are outliers. 

\vspace{4mm}

\textbf{False Positive Classification Rate.} 
Our event classification experiences some false positives. As shown in Figure \ref{fig:windowlevelexample}, we can reduce these false positives through model optimization, including adjustments to architecture, parameter tuning, and the addition of data and features (i.e. adding gyroscope). Currently, there are several events flagged as pickups that are erroneous, and these usually occur in between pickup cycles when the participant is turning around. A naive approach to address this at the event level is to apply a ToP duration filter that removes isolated false positive frames, effectively reducing the false positive rate without compromising recall. In future work, further model optimization can address these false positive classifications using a more appropriate solution. This remains a solvable challenge and a natural future extension of our work.

%% file: writing/discussion.tex

\section{Discussion and Future Works}\label{sec:discussion}

We have approached pickup measurement from a brand-new perspective. Our approach can be further optimized by additional ablation study in the modeling approach and further refinement of the IMU movie design. 

\vspace{2mm}

IMUVIE offers key advantages over frequency-based approaches while sustaining strong performance in ToP timeline action localization and measurement. First, line plots can be less computationally demanding to generate than CWT or FFT plots. Secondly, IMU movies derived from line plots may be more interpretable for humans than either (a) raw IMU numerical data or (b) conventional visualization techniques, such as continuous wavelet transform (CWT) or fast Fourier transform (FFT). We present the timeline activity localization community with a viable alternative to frequency-based methods, with the added benefit of enhancing the interpretability of the input data over CWT and FFT outputs. Line plots, as opposed to frequency plots, are hypothesized to be easier for engineers to link movement data patterns to real-world events, supporting more informed feature design and selection to improve timeline activity classification models. 
In future work, comparing model performance when classifying frequency-based versus line-plot movies will be valuable. We may also consider surveying engineers to assess how they perceive the ease of feature engineering with line plots versus frequency-based features.

\vspace{2mm}


We hypothesize that pre-trained image encoders might not be ideal for our application. Encoders such as I3D~\cite{carreira2017quo} are trained on datasets, such as Kinetics, which include over 400 human action classes optimized for activity classification. While these models excel at recognizing human activities in typical video data, they may not transfer well to our unique IMU movies, where sensor activity is represented as line plots. Future research is needed to determine whether a pre-trained encoder could enhance our model.

\vspace{2mm}

We must also collect additional data on pickup events. This data collection aims to enhance the model’s versatility and improve its performance on new, unseen subjects. Reducing false positive event classifications is crucial, and we have already observed significant improvements by expanding the dataset during model development. Our ultimate goal is to curate a comprehensive dataset of hundreds of seniors performing pickup events and other activities, enabling continuous fall risk monitoring.
Beyond collecting more senior-sourced data, an effective way to strengthen the training set may be to capture additional examples of these movements by having an engineer act out a variety of pickup styles. While not senior-sourced, this augmentation could improve model performance on unusual cases and bolster the dataset. 

%% file: writing/related_works.tex

\vspace{-2mm}


\section{Related Works}\label{sec:related_works}

ToP measurement systems such as IMUVIE and the prior state-of-the-art ToPick~\cite{clapham2024topick} offer an alternative to traditional pressure-sensing approaches used by practitioners to measure pickup ability. Previously, pickup measurement required labor-intensive offline analysis by skilled practitioners. One such existing technology, GAITRite \cite{mcdonough2001validity}, uses pressure mats to record data in clinical settings. ToPick~\cite{clapham2024topick} eliminates the necessity to visit a clinic for ToP measurement by using an AI-based rule-driven model that processes IMU sensor data. ToPick is a mobile healthcare solution~\cite{koltermann2024gait,mandal2017predictive, 9280214,9630392,9630765,10152092,10110718,10112671,9197571,xu2018wistep,hsu2017extracting, rana2016gait,li2018estimation} that can be used anywhere. One key distinction of ToPick is that it is focused on measuring ToP. ToPick could potentially integrate with wearable footwear~\cite{9434861}while surpassing the limitations of 2D pressure data. Using 3D motion data in place of pressure sensors enables the capture of more features that can reliably indicate specific activities, such as a pickup. ToPick pickup measurement leans on IMU sensors that can be used for all kinds of purposes. IMUVIE offers the same function of ToP as ToPick but seeks to use a more generalizable ML approach than a typically overfit rule-based AI model.



\vspace{2mm}

Many works~\cite{koltermann2024gait,6466615, 7173053, mandal2017predictive, 8768050, demrozi2019toward, 9197571, 9280214, 9181210, 9630378, 9630392,10110718, 10112671,pham2018highly,benson2019automated,hori2020inertial,fusca2018method,li2010walking,an2021mgait,keppler2019validity,pillai2020personalized,godfrey2015instrumenting,jasiewicz2006gait,dalton2013analysis,atallah2012validation,mansfield2003use} leverage IMU sensors to record movement data, enabling activity measurement and motion tracking. Some works~\cite{6466615,10134510} even apply IMU data to identify individuals by their gait patterns. More relevant to our approach, IMU sensors have proven useful for monitoring and diagnosing abnormal gait patterns~\cite{xie2022gaittracker}.
Longitudinally, these mobile solutions offer a significant advantage over infrequent clinical testing by allowing frequent, accessible measurements that can be conducted anywhere by untrained users. The limitation of ToPick lies not in its hardware or system design, but in its model, which does not generalize well to new, unseen subjects. IMUVIE achieves the same goal as ToPick of measuring ToP while proving itself superior in its ability to generalize effectively to unseen subjects by not relying on a rule-based model. We achieve a high event level recall of 97\% and consistently accurate ToP detection at the window level, with an average accuracy of 91-92\% over multiple evaluations. 



\vspace{2mm}

Our vision-based approach distinguishes us from ToPick's thresholding state-of-the-art pickup measurement method. 
While we aim to avoid camera sensors, we adopt a similar vision approach to vision-based methods ~\cite{10134510,10152092,9630765,9197571} that employ traditional RGB imaging sensors for tracking or activity recognition. 
Specifically, our pickup localization method draws insights from ActionFormer~\cite{zhang2022actionformer}, a vision model that localizes actions within a traditional video. ActionFormer can be applied to diverse real-world datasets, such as sports videos. We do not use movies with typical RGB camera frames in our scenario. Instead, our movies consist of line plots, presenting a simpler structure than standard RGB videos, such as sports broadcasts. 
This approach competes with the performance of CWT and FFT methods while also reducing some of the complexities inherent in those representations of time-series motion data.
We make several key adaptations to our model design to address the specific requirements of our task. First, because our data is relatively straightforward, we reduce model complexity. Secondly, we choose to omit pre-trained encoders for embedding motion movie frames. Thirdly, we decide that the feature pyramid is not essential for our initial application, though we may explore its impact on performance in future work.
Although we implement these key modifications, we retain the input structure and notation of ActionFormer. Sets \( X \) and \( Y \) represent the IMU movie frames and their respective ground truth labels. The classification for each input frame in \( X \) is denoted by \( \hat{Y} \), while each localized action is described by the tuple \( y_i = (s_i, e_i, a_i) \), which includes the action’s start time \( s_i \), end time \( e_i \), and action label \( a_i \).
IMUVIE integrates methods and benefits from both IMU sensors and vision-based timeline activity localization to create a powerful system capable of generalizable and highly accurate ToP measurements.

%% file: writing/conclusion.tex
\vspace{-3mm}
\section{Conclusion}\label{sec:conclusion}

To enhance the generalizability of pickup measurement beyond the state-of-the-art rule-based approach, we approached ToP measurement as a vision task. We answered our two research questions by presenting the IMUVIE motion movie system and design principles that contribute broadly to timeline action localization across domains. Our IMUVIE model demonstrated strong generalizability and robust performance through multiple rigorous 33-fold, leave one subject out cross validations, achieving an average frame level accuracy range of 91-92\% across multiple rounds of evaluations spanning 256,291 candidate movie frames across the 33 senior study participants. In the event level evaluation, we reached a recall of 97\% across 129 pickup events, effectively demonstrating that most real pickup events are correctly classified by IMUVIE. While further optimizations will improve window level performance and reduce false positives, our work marks significant progress toward a generalizable, accurate, and longitudinally deployable pickup measurement system capable of assessing fall risk and supporting timely expert intervention as movement abilities change. Survey interviews with 23 prospective users indicate positive receptivity among seniors for a wearable pickup ability fall risk monitoring system like IMUVIE.
